\documentclass[letterpaper]{article} 
\usepackage{aaai25}  
\usepackage{times}  
\usepackage{helvet}  
\usepackage{courier}  
\usepackage[hyphens]{url}  
\usepackage{graphicx} 
\usepackage{natbib}  
\usepackage{caption} 
\usepackage{algorithm}
\usepackage{algorithmic}
\usepackage{algorithm}
\usepackage{algorithmic}
\usepackage{amsmath}
\usepackage{amsfonts}
\usepackage{dsfont}
\usepackage{booktabs}
\usepackage{makecell}
\usepackage{pifont}
\usepackage{multirow} 

\usepackage{newfloat}
\usepackage{listings}
\DeclareCaptionStyle{ruled}{labelfont=normalfont,labelsep=colon,strut=off} 
\floatstyle{ruled}
\newfloat{listing}{tb}{lst}{}
\floatname{listing}{Listing}
\nocopyright 

\title{MSSDA: Multi-Sub-Source Adaptation for Diabetic Foot Neuropathy Recognition}
\author{
    Yan Zhong\textsuperscript{\rm 1},
    Zhixin Yan\textsuperscript{\rm 1},
    Yi Xie\textsuperscript{\rm 1},
    Shibin Wu\textsuperscript{\rm 2},
    Huaidong Zhang\textsuperscript{\rm 1},
    Lin Shu \textsuperscript{\rm 1}\thanks{Corresponding author, shul@scut.edu.cn},
    Peiru Zhou\textsuperscript{\rm 3}
}
\affiliations{
    \textsuperscript{\rm 1}South China University of Technology \\
    \textsuperscript{\rm 2}City University of Hong Kong \\
     \textsuperscript{\rm 3}The First Affiliated Hospital of Jinan University    
}

\usepackage{bibentry}

\begin{document}

\maketitle

\begin{abstract}
Diabetic foot neuropathy (DFN) is a critical factor leading to diabetic foot ulcers, which is one of the most common and severe complications of diabetes mellitus (DM) and is associated with high risks of amputation and mortality. Despite its significance, existing datasets do not directly derive from plantar data and lack continuous, long-term foot-specific information. To advance DFN research, we have collected a novel dataset comprising continuous plantar pressure data to recognize diabetic foot neuropathy. This dataset includes data from 94 DM patients with DFN and 41 DM patients without DFN.
 Moreover, traditional methods divide datasets by individuals, potentially leading to significant domain discrepancies in some feature spaces due to the absence of mid-domain data. In this paper, we propose an effective domain adaptation method to address this proplem. We split the dataset based on convolutional feature statistics and select appropriate sub-source domains to enhance efficiency and avoid negative transfer. We then align the distributions of each source and target domain pair in specific feature spaces to minimize the domain gap. Comprehensive results validate the effectiveness of our method on both the newly proposed dataset for DFN recognition and an existing dataset.
\end{abstract}

%

\section{Introduction}

Diabetes mellitus (DM) is a serious non-communicable disease and a global public health issue. According to the International Diabetes Federation \cite{Diabetes693}, approximately 537 million adults worldwide had diabetes in 2021, and this number is projected to reach 693 million by 2045. Over 30\% of diabetic patients will develop diabetic peripheral neuropathy (DPN), with the incidence increasing with age \cite{van2008neuropathy,carls2011economic}. DPN affects the autonomic, sensory, and motor nervous systems, compromising skin integrity and sensation in the feet, thereby increasing susceptibility to injury and diabetic foot ulcers (DFU) \cite{fernando2013biomechanical}.

\begin{figure}[t]
\centerline{\includegraphics[width=\linewidth]{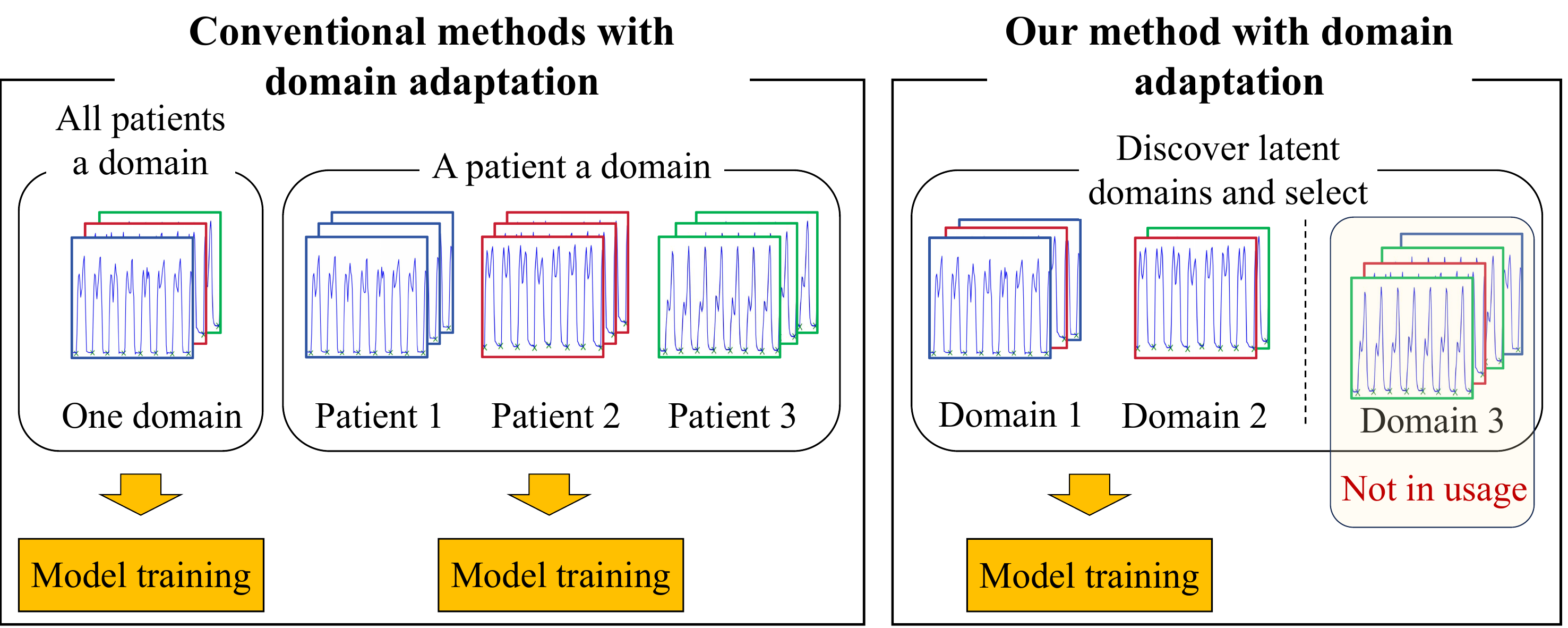}}
\caption{Unlike conventional methods, our method aims to discover latent domains and use the source data with some specific domain labels to train the model but not all of them. }
\label{fig 1}
\end{figure}

Early screening and proactive management can prevent 45\% to 85\% of foot ulcers \cite{american1999consensus}. 
Researches \cite{coppini2001natural,kastenbauer2001prospective} indicate that uneven plantar pressure distribution, particularly high pressure in certain areas, significantly contributes to foot ulcers, with a correlation of 70\% to 90\%. These studies have spurred research into wearable footwear to monitor plantar pressure data  \cite{wang2020novel, sazonov2013development} and analyze real-time changes to screen for peripheral neuropathy early, which is crucial for preventing foot ulcers. 
Studies on plantar pressure abnormality detection have achieved high accuracy in intra-subject setting using artificial neural networks  \cite{wafai}, SVM \cite{botros} and backpropagation neural network \cite{Liuli}.

DFN is a is a specific form of DPN, primarily affecting the nervous system of the feet in diabetic patients. In the field of diabetic foot neuropathy recognition, however, there is no dataset specifically focused on foot information. Some datasets are constructed from electronic health records, including age and pain-related medication prescriptions \cite{dubrava2017using} while others include more complex indicators such as patient medical history, physical examination and biochemical test results \cite{lian2023study}. All these datasets are not directly related to foot information, lacking continuous, long-term foot-specific information.
These issues motivat us to create a dataset for DFN recognition collected using wearable shoes, which contains more foot-specific information. We name it DFN-DS, which includes 5-minute continuous plantar pressure data from 94 DM patients with DFN and 41 DM patients without DFN.

Although many methods perform well under intra-subject settings, they often perform poorly under cross-subject settings due to the significant data distribution gap between individuals. Transfer learning methods are introduced to reduce this gap in biomedical signal processing. In ECG tasks, MS-MDA  \cite{chen2021ms} achieve good performance in emotion recognition. In fall risk assessment tasks, MhNet is proposed based on adversarial domain adaptation, achieving good performance with additional few-step setting \cite{mhnet}.
However, these methods either divide the dataset by subject for multi-source domain adaptation or combine all samples for single-source domain adaptation. We believe the former may fail due to the significant domain gap caused by the absence of mid-domain data to bridge the source and target domain, while the latter does not leverage the advantages of multi-source domain adaptation.

In this paper, we propose a new three-stage alignment framework to overcome these issues. The first stage trains a model to separate all the samples as well as possible using contrastive learning. In the second stage, the original source dataset is divided into K sub-source domains by convolutional feature statistics, where K is determined by the Bayesian Information Criterion. Source samples are then assigned pseudo domain labels. In the third stage, we select the source samples with some proper domain labels according to a strategy to avoid negative transfer, and then we align the distributions of each pair of source and target domains in multiple feature spaces.

The contributions of this paper are summarized as follows: (1) We propose a continuous plantar pressure dataset, the first constructed from plantar pressure data for diabetic foot neuropathy recognition.
(2) We propose a novel framework for biomedical signal processing that divides the dataset by convolutional feature statistics and selects some proper sub-source domains, then aligns the distributions of each pair of source and target domains in multiple feature spaces. (3) We conduct comprehensive experiments on two datasets, validating the effectiveness of the proposed model through experimental results.

\section{Related Work}

\subsection{Single-source Domain Adaptation (SDA)}

SDA aims to reduce the domain gap in the feature space when data may follow different distributions, which often leads to poor performance of traditional methods. Based on the generalization bound \cite{ben2006analysis, ben2010theory} measured by Maximum Mean Discrepancy (MMD), DAN \cite{DAN} is proposed to mitigate the shift in feature space. Deep CORAL \cite{deepCoral} calculate the distribution gap between source domain and target domain using second-order statistics instead of MMD. Subsequently, JAN \cite{JAN} is developed to address the joint distribution gap. MCD \cite{MCD} is then proposed to approximate the disparity difference in the bound by the disagreement between two classifiers' outputs. 
 Domain-adversarial methods are developed with the emergence of Generative Adversarial Network (GAN).
DANN \cite{DANN} is the first to propose adversarial domain adaptation based on the generalization bound and the adversarial idea. 
CDAN \cite{CDAN} advances the theoretical underpinnings with Disparity Discrepancy, pushing the boundaries of domain adaptation methods. Inspired by these methods, MhNet and SFDA are proposed and arrive good performance in falling risk assessment tasks \cite{mhnet, SFDA}.


\subsection{Multi-source Domain Adaptation (MDA)}
Unlike single-source domain adaptation, multi-source domain adaptation involves multiple source domains, introducing more complex inter-domain gaps. DSAN \cite{dsan} divides the dataset into several domains, considering data with the same category labels to share the same domain label. M3SDA \cite{m3fan} aligns multiple source domains with the target domain while also ensuring alignment among all the source domains, which aims to unify data from different domains within a common feature space. In contrast, MFSAN \cite{mfsan} aligns the distributions of each pair of source and target domains in multiple feature spaces, processing N alignments simultaneously for N source domains and a target domain.

\begin{figure*}[!t]
  \centering
  \includegraphics[width=\textwidth]{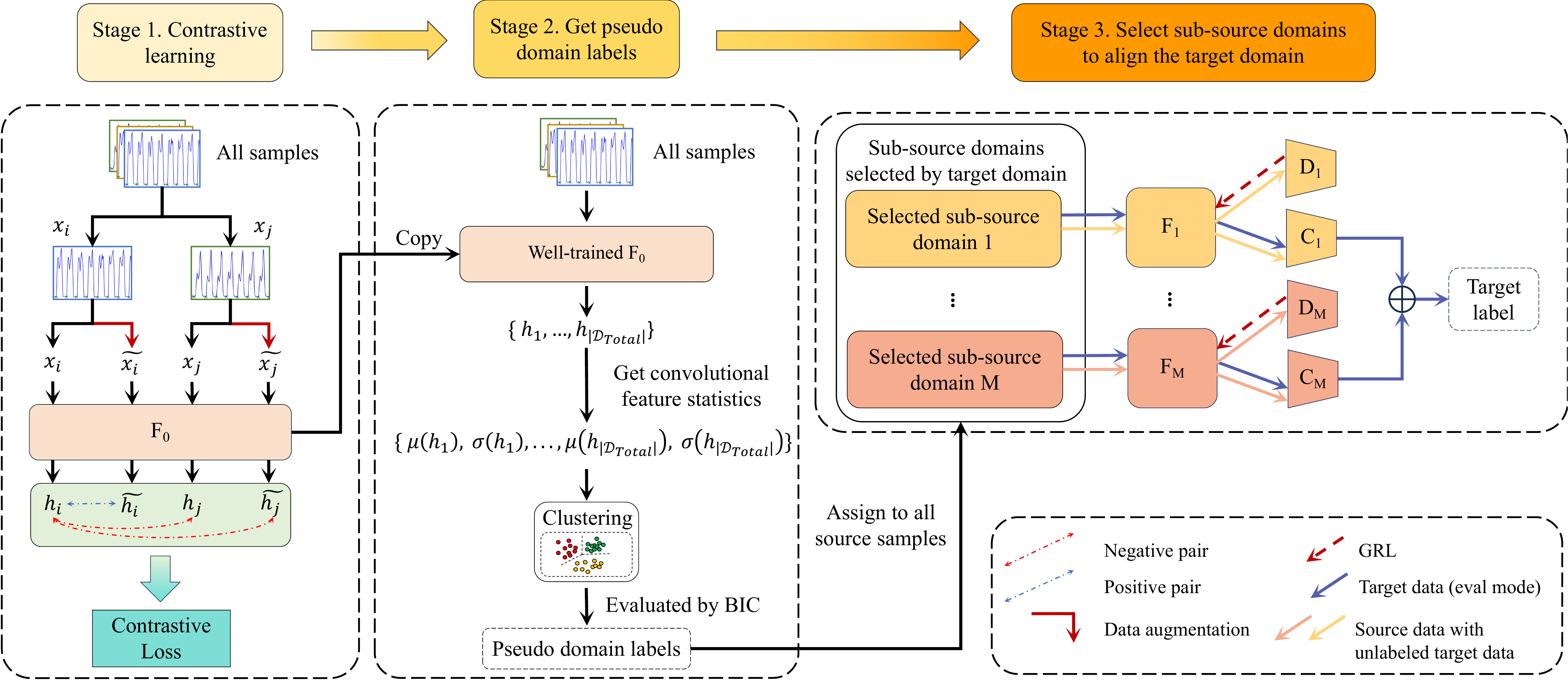} 
  \caption{An overview of the proposed three-stage framework MSSDA. $F_i$ denotes a feature extractor, $D_i$ signifies a domain discriminator, $C_i$ represents a classifier. Note that the $F_0$ trained in stage 1 is used in stage 2 while $F_1$,..., $F_M$ are not fine-tuned from $F_0$. In stage 3, there are specific feature extractors, domain discriminator and classifiers for each selected source domain. Please be aware that $F_0$, $F_1$,..., $F_M$ neither share the network architecture nor the weights. (Best viewed in color.)
  }
\label{fig2}
\end{figure*}

   Many works in the field of biomedical signal process are related to
   MDA methods, such as gait analysis, ECG tasks, EMG tasks and EEG tasks. Usually in their methods, datasets are divided by individual, with each individual’s data constituting a separate domain. Some approaches align multi-source domains and a target domain in a feature spaces using the idea of M3SDA \cite{SFDA, mu2020unsupervised}. Some others process alignment with the idea of MFSAN, meaning that they try to align multi-source domains and a target domain in multi feature spaces \cite{chen2021ms, deng2021multi, she2023multisource, mcdcd}. 
   
   However, these methods are typically designed for datasets with fewer than 50 subjects, most having less than 20. As the number of subjects increases to 100 or even 500, their approaches require substantial computational resources and time due to cross-domain computations and a parallel architecture based on the number of subjects. Moreover, using data from all individuals without selection can lead to negative transfer, while the distribution gap can be hard to reduce because of the absence of mid-domain data to bridge source domain and target domain under a patient a domain setting.
To address these problems, we propose to split datasets based on convolutional feature statistics, rather than relying on subjects as in conventional methods. Additionally, our method incorporates a process for selecting appropriate sub-source domains to avoid negative transfer, ensuring that not all source domains are considered for alignment with the target domain. We then leverage the idea of aligning the distributions of each pair of source and target domains across multiple feature spaces to enhance performance.

\section{Proposed Method}

Considering that we have a source domain constructed by N individuals' data $\mathcal{D}_{S}= {\{\mathcal{D}_{1}, \mathcal{D}_{2},...,\mathcal{D}_{N}\}}$, an target domain constructed by one testing individual's data $\mathcal{D}_{T}$ and total dataset $\mathcal{D}_{Total}=\mathcal{D}_{S} \cup \mathcal{D}_{T}$. Note that $\mathcal{D}_{S} \cap \mathcal{D}_{T}=\emptyset$. We get labeled source samples $\{(x_{i}^{s}, y_{i}^{s})\}_{i=1}^{\left\lvert\mathcal{D}_{S}\right\rvert} $ from $\mathcal{D}_{S}$, where ${\left\lvert\mathcal{D}_{S}\right\rvert} $ refer to the total number of samples in $\mathcal{D}_{S}$. 
Similarly we have unlabelled target samples 
$\{x_{j}^{t}\}_{j=1}^{\left\lvert\mathcal{D}_{T}\right\rvert} $ from target domain $\mathcal{D}_{T}$ and all unlabeled samples 
$\{x_m\}_{m=1}^{\left\lvert\mathcal{D}_{Total}\right\rvert}$ from $\mathcal{D}_{Total}$. Our goal is to mitigate the domain shift between $\mathcal{D}_{S}$ and $\mathcal{D}_{T}$ to learn common domain-invariant features in order to improve prediction accuracy in the target domain.
 
\subsection{Three-Stage Alignment Framework}
Our framework consists of three stages as shown in Figure \ref{fig2}. In the first stage, we try to make all samples distinguishable from each other in the feature space by contrastive learning. 
  Then in the second stage, 
  we feed all samples to the well-trained feature extractor from the first stage and get the outputs,
  then we calculate the convolutional feature statistics (mean and standard deviation) of the output.
  After performing Gaussian Mixture Model (GMM) on the convolutional feature statistics, we obtain pseudo domain labels and assign corresponding ones to all source samples.
  The number of clusters is determined using the Bayesian Information Criterion (BIC). 
 Finally, in the third stage, we select M sub-source domains that are closest to the target domain and align the distributions of each pair of source and target domain in multiple feature spaces.
 
\subsubsection{Stage 1. Contrastive Learning}

Here we show the details in stage 1.
Contrastive learning is widely used as a pre-training method to enhance the feature extractor's ability to learn effective representations in the field of biomedical signal process \cite{commet, natureECG}. The main idea of it is to mine data consistency by bringing similar data (or positive pair) closer together and pushing dissimilar data (or negative pair) further apart. Unlike previous methods, we do not employ contrastive learning as a pre-training technique. Instead, we harness its influence within the feature space to widen the distance between all samples whatever from $\mathcal{D}_{T}$ or $\mathcal{D}_{S}$, thereby enhancing the clustering effect in the second stage, which will be shown in detail below. Specifically, we set $(x_i, \tilde{x}_i)$ as a positive pair, $(x_i, x_j)$ and $(x_i, \tilde{x}_j)$ as negative pairs, where $\tilde{x}_i$ is the augmented view of $x_i$, $i \neq j$ and $x_i, x_j \in \mathcal{D}_{Total}$. We then make all samples distinguishable from each other in the feature space using contrastive loss, defined as

\begin{equation}
\begin{aligned}
    \mathcal{L_C} =  \mathbb{E}
\left[ -\log \frac{\exp(h_i \cdot \tilde{h_i})}{\sum_{j=1}^J \exp(h_i \cdot \tilde{h}_j) + \mathds{1}_{[i \neq j]} \exp(h_i \cdot h_j)} \right],
\end{aligned}
\label{stage1.1}
\end{equation}
    where $h_i=F_{0}(x_i), {x_i \in \mathcal{D}_{Total}}$, $J={\left\lvert\mathcal{D}_{Total}\right\rvert} $, $F_0$ stands for the feature extractor used in this stage, $\cdot$ means dot product and $\mathds{1}_{[i \neq j]}$ stands for an indicator function that equals 1 when $i\neq j$ and 0 other wise. Note that the feature extractor $F_0$ well-trained in this stage will be used in following stages.

\subsubsection{Stage 2. Clustering and Get Pseudo Domain Labels}
Inspired by Matsuura and Harada's work \cite{discoverlatentdomains},
 we assume that the latent domains of data are reflected in their styles, specifically in the convolutional feature statistics (mean and standard deviations). Therefore, we firstly obtain convolutional feature statistics from the well-trained feature extractor $F_0$.  Specifically, we calculate the mean and standard deviations of the source samples by channel c,
 
\begin{equation}
\begin{aligned}
\mu_{c}(F_{0}(x_{i})) &= \frac{1}{HW}\sum_{h=1}^{H}\sum_{w=1}^{W} (F_{0}(x_{i}))_{chw}, \\
\label{stage2.1}
\end{aligned}
\end{equation}

\begin{equation}
\begin{aligned}
\sigma_{c}(F_{0}(x_{i})) = \sqrt{\frac{1}{HW}\sum_{h=1}^{H}\sum_{w=1}^{W} (F_{0}(x_{i})_{chw}-\mu_{c}(F_{0}(x_{i})))^2 },\\
\label{stage2.2}
\end{aligned}
\end{equation}
where $x_{i} \in \mathcal{D}_{Total}$, $c, h, w$ respectively refer to channel, height and width of the representation of $x_{i}$ transformed by well-trained $F_0$ in the feature space.

We represent them in a simple concise form in Figure 2, in which 
 $h_i=F_{0}(x_{i})$, 
 $\mu(h_i)=\{\mu_{c}(F_{0}(x_{i})\}_{c=1}^{C}$, 
 $\sigma(h_i)=\{\sigma_{c}(F_{0}(x_{i})\}_{c=1}^{C}$.
Then we utilize GMM as the clustering method on these convolutional feature statistics. Finally, the influence of different cluster numbers will be evaluated by the Bayesian Information Criterion (BIC) \cite{schwarz1978estimating}, motivated by SSDA \cite{lu2021cross}, 

\begin{equation}
\begin{aligned}
BIC = \text{-} 2\ln{L}+k\ln{m},
\end{aligned}
\label{stage2.3}
\end{equation}
 where $L$ represents the maximized value of the likelihood function for the estimated model, $k$ represents the number of free parameters to be estimated, and $m$ is the sample size. We seek proper cluster number K which minimizes BIC. 
 
 Using BIC, we determine a certain number of clusters, which differs from Matsuura and Harada's approach. Additionally, they utilize a stack of convolutional feature statistics obtained from lower layers of the feature extractor, whereas we choose those from the last layer because it is task-specific that separate the data as effectively as possible, aligning with our objectives.
 
  In the end of stage 2, after assigning K types of numerical pseudo-domain labels to source samples, we get K sub-source domains and  K cluster centers, denoted as $\mathcal D_{sub_k}$  $,Center_{sub_k}, k \in \{ 1,...,K\}$, respectively.

 \subsubsection{Stage 3. Select Sub-Source Domains to Align the Target Domain}
 Compared with the previous works in the field of biomedical signal process, our method do not use all the source domains, instead, we select some of them to avoid negative transfer which may be happened as the individual number of the dataset increases.
 We select $M$ sub-source domains by the distance between the sub-source domain center $Center_{sub_k}$ and the target samples, which is calculated as followed:

\begin{equation}
\begin{aligned}
dis_{k} = \max_{x_{j}^{t} \sim \mathcal{D}_{T}} \left\|   T(\mu(F_{0}(x_{j}^{t})), \sigma(F_{0}(x_{j}^{t}))) , Center_{sub_k}  \right\|_2,
\end{aligned}
\label{stage3.1}
\end{equation}
where $T$ stands for PCA with dim = 2 and $\| \cdot \|_2$ refers to $L_2$ norm. Then we select $M$ ($M \leq K$) sub-source domains by the distance calculated in Eq.(\ref{stage3.1}) :

\begin{equation}
\begin{aligned}
\mathcal D_{ssub_1} = \operatorname*{argmin}_{k \in \{ 1, \ldots, K \}} dis_k,
\label{stage3.2}
\end{aligned}
\end{equation}

\begin{equation}
\begin{aligned}
\mathcal D_{ssub_M} = \operatorname*{argmin}_{k \in \{ 1, \ldots, K \} \setminus \{\mathcal D_{ssub_1},...,\mathcal D_{ssub_{M-1}} \} }  dis_k,
\label{stage3.3}
\end{aligned}
\end{equation}
where $M$ is determined by user. We name them selected sub-source domain 1,..., selected sub-source domain M, respectively, as shown in Figure 2.
Then we try to reduce the domain gap between these $M$ sub-source domains and target domain in order to improve cross-subjects performance. 
We utilize the idea  that aligns the distributions of each pair of source and target domains in multiple feature spaces \cite{mfsan}. In a traditional domain-adversarial framework, such as DANN \cite{DANN}, there are three parts, a feature extractor $F$ to get domain-invariant features, a domain discriminator $D$ to distinguish which domain the output of the feature extractor is from and a classifier $C$ to predict the category label of the output of the feature extractor. By loss function, $F$ is encouraged to extract the features that are challenging for $D$ to distinguish, while the $D$ is trained to correctly predict the domain label of the output of the feature extractor. This creates an adversarial relationship between $F$ and $D$, allowing $F$ to extract domain-invariant features across different domains. The framework we used in stage 3 can be seen as a parallel architecture of $M$ DANNs. The target domain is aligned with each source domain in a specified sub-networks, as showed in Figure 2. Loss function is calculated as bellow:

\begin{equation}
\begin{aligned}
    \mathcal{L}_{\text{cls}} = &  \sum_{k=1}^{M}\sum_{(x_{i}^{s},y_{i}^{s}) \sim \mathcal D_{ssub_k}} \mathcal{L}_{CE}(C_k(F_k(x_{i}^{s})),y_{i}^{s}),
\label{stage3.4}
\end{aligned}
\end{equation}

\begin{equation}
\begin{split}
\begin{aligned}
    \mathcal{L}_{\text{adv}} =   \sum_{k=1}^{M}\sum_{\substack{x_{i}^{s} \sim \mathcal {D}_{ssub_k}, \\ x_{j}^{t}  \sim \mathcal{D}_T}} 
    &\mathcal{L}_{CE}(D_k(F_k(x_{i}^{s})), 1)+\\
    &  \mathcal{L}_{CE}(D_k(F_k(x_{j}^{t})), 0),
\label{stage3.5}
\end{aligned}
\end{split}
\end{equation}

\begin{equation}
\begin{split}
\begin{aligned}
    \mathcal{L}_{\text{total}}=\mathcal{L}_{\text{cls}}-\alpha \mathcal{L}_{\text{adv}},
\label{stage3.6}
\end{aligned}
\end{split}
\end{equation}
where $\mathcal{L}_{\text{CE}}$ denotes the use of cross-entropy,  $\mathcal{L}_{\text{cls}}$ represents the category classification loss, $\mathcal{L}_{\text{adv}}$ refers to the adversarial loss, $\alpha$ is a trade-off parameter and $\mathcal{L}_{\text{total}}$ is the total loss. 
Finally, to predict the labels of target samples, we compute the average of all classifier outputs.

\begin{algorithm}[h]
\caption{Training algorithm}
\label{alg:algorithm}
\textbf{Require}: source samples $\{(x_{i}^{s}, y_{i}^{s})\}_{i=1}^{|\mathcal{D}_{S}|}$, target data $\{x_{j}^{t}\}_{j=1}^{|\mathcal{D}_{T}|}$, trade-off parameter $\alpha$ and batch size $bs$
\begin{algorithmic}[1] 
\STATE Initialize parameters of $F_0$, $F_1$, $F_2$, $D_1$, $D_2$, $C_1$, $C_2$
\STATE Train $F_0$ by Eq.(\ref{stage1.1}) using all samples 
$
\left\{ x_{m} \right\}_{m=1}^{|\mathcal{D}_{Total}|} = 
\left\{ x_{i}^{s} \right\}_{i=1}^{|\mathcal{D}_{S}|} 
\cup 
\left\{x_{j}^{t} \right\}_{j=1}^{|\mathcal{D}_{T}|}$

\STATE Collect the representation $\{F_0(x_{m})\}_{m=1}^{|\mathcal{D}_{Total}|}$ of all the samples and get their convolutional feature statistics $\{ \mu(h_m), \sigma(h_m) \}_{m=1}^{|\mathcal{D}_{Total}|}$ by Eq.(\ref{stage2.1}) (\ref{stage2.2}), then use GMM as cluster method on these convolutional feature statistics and get the proper cluster number $K$ evaluated by BIC
\STATE Assign pseudo domain labels of $K$ types to all source samples 
\STATE Select $M$ ($M \leq K$) proper sub-source domains $\mathcal D_{ssub_M}$,..., $\mathcal D_{ssub_M}$ by Eq.(\ref{stage3.2}) (\ref{stage3.3})
\REPEAT
\FOR{ $k=1$ to $M$}
\STATE Sample $bs$ samples $\{x_{i}^{ssub_k}\}_{i=1}^{bs}$ from $\mathcal D_{ssub_k}$ 
\STATE Sample $bs$ samples $\{x_{j}^t\}_{j=1}^{bs}$ from $\mathcal{D}_{T}$
\STATE Feed $x_{i}^{ssub_k}$ and $x_{j}^{t}$ into $F_k$ and get $F_k(x_{i}^{ssub_k})$, $F_k(x_{j}^{t})$
\STATE Feed $F_k(x_{i}^{ssub_k})$ into $C_k$ and get $C_k(F_k(x_{i}^{ssub_k}))$ 
\STATE Feed $F_k(x_{i}^{ssub_k})$, $F_k(x_{j}^t)$ into $D_k$ and get $D_k(F_k(x_{i}^{ssub_k}))$, $D_k(F_k(x_{j}^t))$
\STATE Calculate $\mathcal{L}_{\text{cls}}$ using $C_k(F_k(x_{i}^{ssub_k}))$ by Eq.(\ref{stage3.4})
\STATE Calculate $\mathcal{L}_{\text{adv}}$ using $D_k(F_k(x_{i}^{ssub_k}))$, $D_k(F_k(x_{j}^t))$ by Eq.(\ref{stage3.5})
\STATE Calculate $\mathcal{L}_{\text{total}}$ by Eq.(\ref{stage3.6})
\STATE Update $F_k$, $C_k$, $D_k$ to minimize $\mathcal{L}_{\text{total}}$
\ENDFOR
\UNTIL{convergence}
\end{algorithmic}
\end{algorithm}

\section{Experiments}

We evaluate our proposed method (MSSDA) with some transfer learning methods that are widely used in the field of biomedical signal process on two datasets: one is our proposed dataset DFN-DS and the other one is Fall Risk Assessment dataset \cite{hu2022dataset}, both involving plantar pressure data.  
By using these two dataset, we verify whether our method is suitable for biomedical signal datasets.
Our code and dataset DFN-DS will be available.
\subsection{Data Preparation}
\subsubsection{Diabetic Foot Neuropathy Dataset (DFN-DS)}
It includes plantar pressure data from 135 subjects, with 94 diabetic patients having DFN (labeled as 1) and 41 without DFN (labeled as 0). We recorded data as each patient walked freely in a straight line for 5 minutes. After data cleaning  we obtained 6,983 samples. Data has the shape $x \in \mathbb R^{147 \times 16}$, where 147 is the time length and 16 is the channel number. In the intra-subject setting, classic methods like  Random Forest, XGBoost, LSTM and 1D-CNN network achieved the accuracy of at least 98\%. More details are in the supplement. All experiments use leave-one-subject-out cross-validation (LOSO-CV) with a common vote threshold of 50\%. In this setup, each subject acts as the target domain $\mathcal{D}_T$, while the rest are the source domain $\mathcal{D}_S$ rotating through all patients.

\subsubsection{Falling Risk Assessment Dataset (FRA) \cite{hu2022dataset} } 
It is a plantar pressure dataset comprises 48 subjects and 7,462 samples, with 23 high-risk subjects (labeled as 1) and 25 low-risk subjects (labeled as 0). The plantar pressure data has the shape $x \in \mathbb R^{69 \times 16}$, where 69 is the time length and 16 is the channel number. All experiments are also under the setting of LOSO-CV. Additionally, experiments on FRA follow a 11-step setting with a 50\% threshold as \cite{mhnet}. That indicates we split all target data into several segments, each containing 11 neighboring samples. If 50\% or more of the samples in a segment are classified as high risk, the entire segment is considered high risk. Performance is calculated by segments, not subjects. This challenging setup closely mimics real-world scenarios.

\subsection{Baselines and Implementation Details}

\subsubsection{Baselines}
We compare MSSDA with various kinds of transfer learning methods that are wildly used in biomedical signal process, including Deep CORAL (D-CORAL) \cite{deepCoral}, Deep Adaptation Network (DAN) \cite{DAN}, Joint Adaptation Network (JAN) \cite{JAN}, Maximum Classifier Discrepancy (MCD) \cite{MCD}, Domain-Adversarial Neural Network (DANN) \cite{DANN}, Conditional Adversarial Domain Adaptation (CDAN) \cite{CDAN} and Multi-scale spatio-temporal hierarchical network (MhNet) \cite{mhnet}. Note that ERM refers to Empirical Risk Minimization, which means to train the model without transfer learning loss.

\subsubsection{Implementation Details}
All methods are implemented using the PyTorch framework and reproduced on a GeForce GTX 4060. In Stage 1, we utilize a network with 4 layers of 1D CNN for contrastive learning on DFN-DS, and a network with 3 layers of 1D CNN for the FRA. In Stage 3, the feature extractor for DFN-DS remains consistent across all methods, comprising 7 layers of 1D CNN, while the FRA uses 3 layers of 1D CNN. Additionally, the domain discriminator and classifier frameworks in the domain-invariant methods are identical, featuring 3-layer fully connected networks for DFN-DS and 2-layer fully connected networks for FRA. All experiments are conducted after balancing the dataset using data reuse techniques.

While in stage 1, we train the feature extractor $F_0$ by contrastive learning for 5000 epochs using AdamW as the optimizer with initial learning rate of 5e-3, a batch size of 64 for DFN-DS. As for FRA, we set the initial learning rate as 1e-3 with  a batch size of 32, other paremeters remain the same.
In stage 3, same as other methods, we use Adam as our optimizer with a weight decay set to 1e-4. We select the learning rate $lr$ from \{1e-2, 8e-3, 5e-3\} for best performance. Additionally, we select the weights of the  domain adaptation loss from \{0.2, 0.5, 1, 2\} to get the best performance of the models, which is 1 for our method in FRA and 2 on DFN-DS. For both DFN-DS and FRA, $M$ used in our method is set to 2.

\begin{table}[!htbp]
\centering
\caption{Performance Comparison of Classification on DFN-DS.} 
\label{tab:tableTab} 
\begin{tabular}{ccccc} 
\toprule 
Method & Precision & Recall & Accuracy & F1  \\
\midrule 
ERM     & 0.767 & 0.596 & 0.593 & 0.671 \\
D-CORAL & 0.731 & 0.840 & 0.674 & 0.782 \\
DAN     & 0.851 & 0.914 & \underline{0.830} & \underline{0.882} \\
JAN     & 0.844 & 0.809 & 0.763 & 0.826 \\
MCD     & \underline{0.908} & 0.840 & \underline{0.830} & 0.873 \\
DANN    & 0.796	&0.829	&0.733	&0.813 \\
CDAN    & 0.767 & \textbf{0.979} & 0.778 & 0.860 \\
MhNet	&0.835   &0.702	&0.696	&0.763\\
Ours    & \textbf{0.916} & \underline{0.926} & \textbf{0.889} & \textbf{0.921} \\
\bottomrule 
\end{tabular}%
\end{table}

\begin{table}[!htbp]
\centering
\caption{Performance Comparison of Classification on FRA.} 
\label{bbb} 

\begin{tabular}{ccccc} 
\toprule 
Method & Precision & Recall & Accuracy & F1 \\
\midrule 
ERM     & 0.491 & 0.717 & 0.696 & 0.583 \\
D-CORAL & 0.615 & 0.645 & 0.642 & 0.630 \\
DAN     & 0.618 & 0.625 & 0.663 & 0.622 \\
JAN     & 0.542 & 0.548 & 0.591 & 0.545 \\
MCD     & 0.653 & 0.749 & 0.723 & 0.698 \\
DANN    & 0.633 & 0.530 & 0.628 & 0.577 \\
CDAN    & 0.583 & 0.647 & 0.707 & 0.613 \\
MhNet   & \underline{0.697} & \underline{0.764} & \textbf{0.730} &\underline{0.729} \\
Ours    & \textbf{0.710} & \textbf{0.796} &\underline{0.729} & \textbf{0.750} \\
\bottomrule 
\end{tabular}%

\end{table}

\subsection{Comparison Results}

\subsubsection{Results on DFN-DS} As shown in Table 1, comparing all methods on our proposed dataset DFN-DS, our approach excels across almost all metrics. In terms of recall, while our method ranks second, it still demonstrates substantial improvement compared to alternative approaches. Most importantly, our method achieves an accuracy that is at least 5.9\% higher than other methods, which is quite remarkable. Note that the high precision of our method indicates that few patients without DFN are mistakenly predicted as having DFN. Specifically, the cluster number evaluated by BIC is 6 on DFN-DS.

\subsubsection{Results on FRA} As shown in Table 2, under challenging and real-time conditions, our approach excels across almost all metrics.  We achieve the highest recall at the cost of a 0.001\% lower accuracy compared to MhNet. Note that we label gaits with high fall risk as positive cases. Thus, recall is a critical metric in our context, as it emphasizes our ability to identify individuals at high risk of falling, which is significantly more important in real-world scenarios. Specifically, the number of clusters evaluated by BIC is 11 on FRA.

\section{Further Analysis}
\subsection{Ablation Study}
In this section, we describe ablation studies to investigate
the effect of different components of our method with $M=2$ on DFN-DS.

\begin{table}[!htbp]
\centering
\caption{Results of the ablation study on stage 3 of our method on DFN-DS.} 
\label{bbb} 
\begin{tabular}{ccccccc} 
\toprule 
\makecell[c]{SA} &
\makecell[c]{MA} & 
\makecell[c]{Select} & 
Prec & Recall & Acc & \makecell[c]{F1} \\
\midrule
\quad & \quad & \quad & 0.767 & 0.596 & 0.593 & 0.670 \\

\ding{52} & \quad & \quad &0.796 &0.829 &0.733 &0.813\\

\quad & \ding{52} & \quad & 0.829 & \textbf{0.926} & 0.815 & 0.874 \\

\ding{52} & \quad & \ding{52} & 0.850 & 0.904 & 0.822 & 0.876 \\

\quad & \ding{52} & \ding{52} & \textbf{0.916} & \textbf{0.926} & \textbf{0.889} & \textbf{0.921} \\

\bottomrule 
\end{tabular}%
\end{table}
In Table 3, we analyze the influence of aligning
the distributions of each pair of source and target domains in multiple feature spaces (MA), selecting proper sub-source domains as stage 3 in proposed method (Select) and  mixing the data from all sub-source domains as a source domain and applying alignment in a single feature space, actually the application of DANN (SA). It is important to note that MA and SA are mutually exclusive processes. 
From the results shown in Table 3, we have the following insightful observations:
\begin{itemize}

\item Note that the one with MA and Select is actually our proposed method. It outperforms all other methods across all metrics, demonstrating the effectiveness of each stage.

\item Perform alignment in several feature spaces (MA) overcome the alignment in a single feature space (SA), whether after sub-source domain selection or not.

\item The idea that aligning a source domain and target domain in a feature space (MA) helps the model to get a better performance on recall.

\item The proposed idea that we should carefully select proper domains (select) to avoid negative transfer is proved, whatever perform the alignment in several feature spaces (MA) or merely in a single feature space (SA).

\end{itemize}

\begin{table}[!htbp]
\centering
\caption{Results of the ablation study on stage 1 and stage 2 of our method on DFN-DS.} 
\label{bbb} 

\begin{tabular}{ccccc} 
\toprule 
GMM & \makecell[c]{ Statistics} & \makecell[c]{CL} & \makecell[c]{Cluster  number} & Acc\\
\midrule 
\ding{52} & \quad & \quad & 9  &0.770\\
\ding{52} & \ding{52} & \quad & 23  &0.815\\
\ding{52} & \quad   &\ding{52} &7   &0.785\\
\ding{52} & \ding{52} & \ding{52} & 6  &0.889\\
\bottomrule 
\end{tabular}%

\end{table}

In Table 4, we analyze the influence of convolutional feature statistics (Statistics) and contrastive learning (CL) used in stages 1 and 2 of our method on DFN-DS. All experiments include sub-source selection in stage 3. Each component contributes to the improvement of the final accuracy. Additionally, we observe the effectiveness of contrastive learning as a strategy for expanding the distances between samples. After processing with the contrastive learning module, the data structure becomes clearer and more separable, enabling GMM to identify fewer but more representative clusters.

\begin{table}[h]
\centering
\caption{Results of the performance using different dataset partitioning method on DFN-DS.} 
\label{bbb} 
\begin{tabular}{cccc} 
\toprule 
\makecell[c]{Method } & \makecell[c]{Specificity }  & \makecell[c]{Recall} &Acc\\
\midrule 
MCDCD	&0.00	&\textbf{1.00}  &0.696\\
MS-MDA	&0.00	&\textbf{1.00}   &0.696\\
DSAN    &0.122  &0.936 &0.689\\
MS-DANN	&0.561	&0.926  &0.815\\ 
Ours    &\textbf{0.805 } &0.926 &\textbf{0.889}\\
\bottomrule 
\end{tabular}%
\end{table}

\subsection{Other Strategy to Divide Source Domain}
In Table 5, we compare our method of dividing the source domain by convolutional feature statistics with other conventional MDA methods. MS-MDA \cite{chen2021ms} and MCDCD \cite{mcdcd} split the source dataset by individuals, while DSAN \cite{dsan} splits the source dataset by labels. MS-DANN, which corresponds to the method with MA only in Table 3, can be seen as a variant of MS-MDA that uses our proposed dataset partitioning method.
 We found that MCDCD and MFSAN are sensitive to the target domain with label 1 but not label 0, leading to performance drops during training. This issue persists even with balanced datasets.
 We believe this is due to the a patient a domain setting, which results the lack of diversity in each branch network and leads to a more extreme lack of diversity in information compared to DSAN. 
 This also leads to a shortage of mid-domain samples to bridge the source and target domains and mitigate the significant domain gap, a problem addressed and partially solved by PMTrans \cite{zhu2023patch}. This highlights the effectiveness of the dataset division method proposed in this paper.

\begin{table}[!htbp]
\centering
\caption{Results of the performance using different strategies to select sub-source domain(s) as source domains on DFN-DS.} 
\label{bbb} 
\small
\begin{tabular}{ccccc} 
\toprule 
\makecell[c]{Strategy} & 
Precision & Recall & Acc & \makecell[c]{F1} \\
\midrule 
\makecell[c]{Top 1 min dis.} & 0.872 & 0.872 & 0.822 & 0.872  \\
\makecell[c]{Top 2 min dis.(Ours)} & \textbf{0.916} & 0.926 & \textbf{0.889} & \textbf{0.921}  \\
\makecell[c]{Top 3 min dis.} & 0.862 & 0.904 & 0.837 & 0.885  \\
\makecell[c]{Top 1 min sum} & 0.821 & 0.926 & 0.807 & 0.870 \\
\makecell[c]{Top 2 min sum} & 0.822 & 0.936 & 0.815 & 0.875 \\
\makecell[c]{Top 3 min sum} & 0.840 & \textbf{0.947} & 0.837 & 0.890 \\
\makecell[c]{All in usage} &0.829	&0.926	&0.815	&0.874\\
\bottomrule 
\end{tabular}%
\end{table}

\subsection{Other Strategies to Select Sub-Source Domains}
In Table 6, dis. stands for the distance calculated in Eq. (5)  while sum refers to the total sum of  Euclidean distances between all target samples and the cluster centers. 
All in usage refers that we use all sub-source domains to train the model, which corresponds to
the method with MA only in Table 3. Considering comprehensively, our strategy is the best according to the results.

\begin{table}[h]
\centering
\small
\caption{Results of the performance using different sub-source domains as source domain on DFN-DS.} 
\label{bbb} 
\begin{tabular}{cccc} 
\toprule 

\makecell[c]{Sub-source domain selection } & \makecell[c]{Specificity}  & \makecell[c]{Recall} &Acc\\
\midrule
 mixed (DANN)    &0.512  &0.829  &0.733\\ \hline
Source = $\mathcal D_{sub_1}$ &0.512	&\textbf{0.968}  &0.830\\
Source = $\mathcal D_{sub_2}$ &0.537	&0.872  &0.770\\
Source = $\mathcal D_{sub_3}$ &0.585	&0.947  &0.837\\
Source = $\mathcal D_{sub_4}$ &\textbf{0.927}	&0.755  &0.807\\
Source = $\mathcal D_{sub_5}$ &0.561	&0.947  &0.830\\
Source = $\mathcal D_{sub_6}$ &0.683	&0.862  &0.807\\ 
 \hline
\makecell[c]{Source = $\mathcal D_{sub_1}$ and $\mathcal D_{sub_4}$} &0.659	&0.926 &0.844\\ 
Top 2 min dis.(Ours)    &0.805	&0.926  &0.889\\
\bottomrule 
\end{tabular}%
\end{table}

\subsection{Sub-source Domain Performance}
In Table 7, we firstly compare the performance of using different single sub-source domain as the source domain on DANN versus using the mixed source domain on DANN with DFN-DS. The results show that expect $\mathcal D_{sub_4}$, all other sub-source domain outperform DANN on all metrics. $\mathcal D_{sub_4}$, however, exhibits significantly better performance on negative samples. These experiments reveal that not all source samples are beneficial for alignment. In a data-driven manner, we set  $\mathcal D_{sub_1}$ and  $\mathcal D_{sub_4}$ as the selected sub-source domains used in stage 3. However, the performance of this setup is worse than our proposed approach.

\begin{figure}[!htbp]
\centerline{\includegraphics[width=\linewidth]{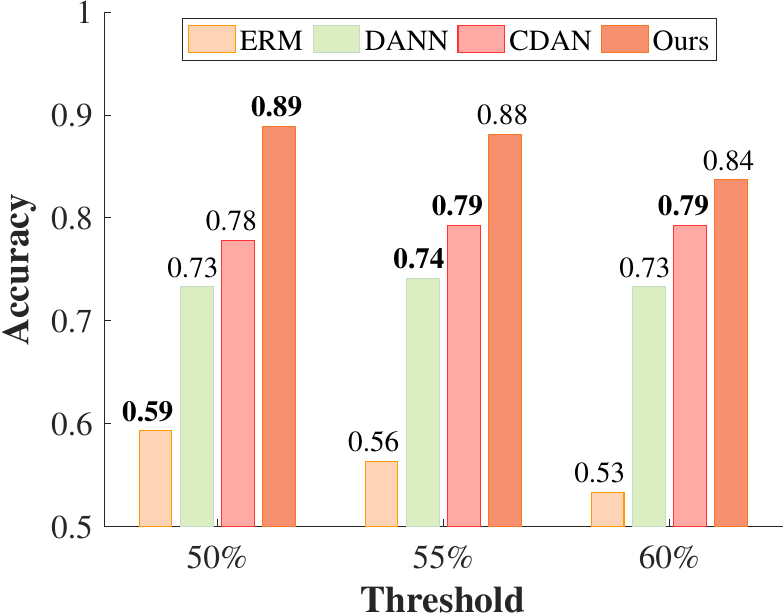}}
\caption{Results of the accuracy using other thresholds on
DFN-DS. }
\label{fig 3}
\end{figure}

\subsection{Performance in Other Thresholds}
In Figure \ref{fig 3}, we compare the accuracy of our method with other domain-adversarial methods at different thresholds.
These transfer learning methods improve accuracy on the target domain, but our method's accuracy drops more significantly as the threshold increases, likely due to excessive pursuit of transferability \cite{chen2019transferability,cui2022discriminability,cui2020towards}. This issue poses a challenge to the credibility of our approach and will be a focus of our future work.

\section{Conlusion}
We propose a dataset for Diabetic Foot Neuropathy (DFN) recognition, which includes continuous plantar pressure data from 94 DM patients with DFN and 41 DM patients without DFN. Previous works in biomedical signal processing either divide the dataset by patient or do not separate the dataset at all. Additionally, few of these studies carefully select data to avoid negative transfer. Our framework addresses these shortcomings. It divides the dataset based on convolutional feature statistics and employs a straightforward yet effective strategy to select appropriate sub-source domains for multi-source domain adaptation, simultaneously aligning the domain-specific distributions of each source-target domain pair. Extensive experiments on two plantar pressure datasets demonstrate the effectiveness of the proposed framework.

\clearpage

\end{document}